\documentclass[10 pt, conference]{ieeeconf}
\IEEEoverridecommandlockouts
\usepackage{graphics}
\usepackage{graphicx}
\usepackage{subfigure}
\usepackage{amsmath}
\usepackage{amssymb}
\usepackage{mathtools}
\usepackage{url}
\usepackage{booktabs}
\usepackage[font=small]{caption}
\usepackage{multirow}
\usepackage{afterpage}
\usepackage{url}
\usepackage{balance}
\usepackage{wasysym}
\usepackage{color}
\usepackage{threeparttable}

\title{\LARGE \bf Compact $\mathbf{3}$D Map-Based Monocular Localization Using Semantic\\ Edge Alignment}
\author{Kejie Qiu, Shenzhou Chen, Jiahui Zhang, Rui Huang, Le Cui, Siyu Zhu, and Ping Tan 
  \thanks{
  All the authors are with Alibaba A.I. Labs, Hangzhou, China.
  {\tt\small kejie.qkj@alibaba-inc.com}.}
  }

\begin{document}
\maketitle 

\begin{abstract}
Accurate localization is fundamental to a variety of applications, such as navigation, robotics, 
autonomous driving, and Augmented Reality (AR). Different from incremental localization,
global localization has no drift caused by error accumulation, which is
desired in many application scenarios. 
In addition to GPS used in the open air, $3$D maps are also widely used 
as alternative global localization references.
In this paper, we propose a compact $3$D map-based global localization
system using a low-cost monocular camera and an IMU (Inertial Measurement Unit).
The proposed compact map consists of two types of simplified elements
with multiple semantic labels, which is well adaptive to various man-made
environments like urban environments. Also, semantic edge features
are used for the key image-map
registration, which is robust against occlusion and long-term appearance changes
in the environments. To further improve the localization performance, 
the key semantic edge alignment is formulated as an optimization problem
based on initial poses predicted by an independent VIO (Visual-Inertial Odometry) module.
The localization system is realized with modular design in real time.
We evaluate the localization accuracy through real-world experimental results
compared with ground truth, long-term localization performance is also
demonstrated.
\end{abstract}

\section{Introduction}
With the fast development of large-scale $3$D reconstruction and high-level autonomous driving,
various $3$D map productions have become widely available nowadays. Given a $3$D map as
the global localization reference, $6$-DoF sensor pose (position and orientation) can be
derived using specific pose estimation algorithms. Map-based localization is essentially
a registration problem to match the sensor observations with a prior map.
Usually high-quality sensors such as GPS-RTK (real-time kinematic), 
Lidar, and SINS (Strapdown Inertial Navigation System)
are used for High Definition (HD) map construction,
and low-cost sensors are used for localization.
For example, a point cloud map constructed by a Lidar can not only be used for Lidar-based
localization, but also a monocular camera-based localization approach \cite{wolcott2014visual}.
A more general vision-based localization method is to utilize a map consists of a mass of visual features
and their descriptors \cite{burki2016appearance}. However, this map is difficult to maintain 
and the map size is too large for large-scale environments.

\begin{figure}[t]
\begin{center}
\includegraphics[width=0.8\columnwidth]{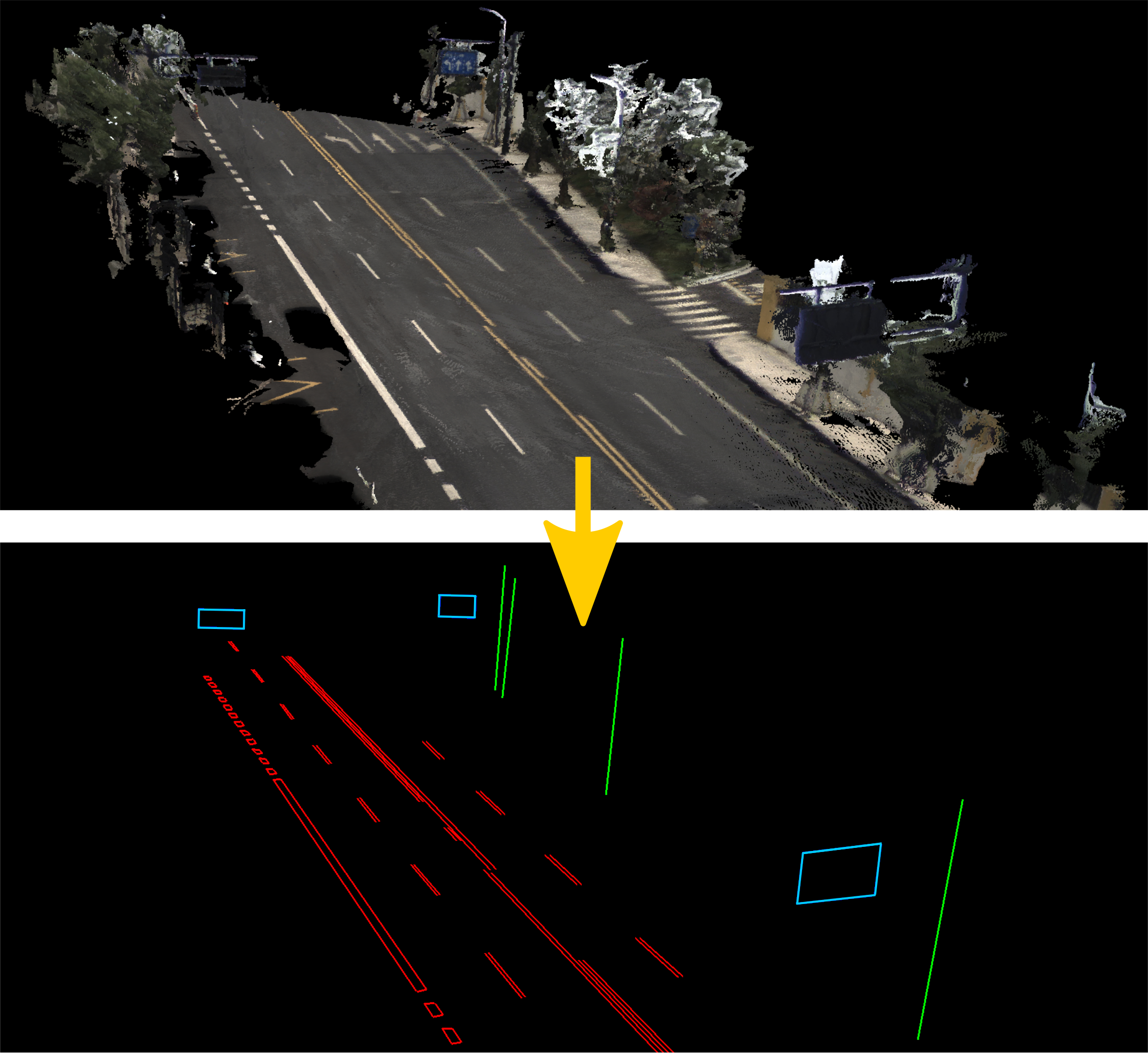}
\caption{Map compaction using the proposed compact map format in an urban environment.
The map size is significantly reduced while the key landmarks are reserved using
two types of line segments (a line segment \& a wireframe). Each type can be
labeled with multiple semantic categories.}
\label{fig:map_compression}
\end{center}
\vspace{-5mm}
\end{figure}

Besides point cloud maps and visual feature maps, there are many other map formats such as textured mesh
model constructed by visual reconstruction, and even vector maps.
Recently, the growing popularity of vector map standards such as OpenDrive \footnote{http://opendrive.org}
and NDS \footnote{http://nds-association.org/} indicates that a
compressed map is a tendency for a light-weight and scalable map representation.
To make full use of available $3$D maps, we propose to
use a more compact map that only consists of semantic line segments and wireframes. Because
plenty of structural elements exist in man-made environments and many of them are comprised
of straight line segments, such as lane lines, lamp poles, traffic signs in outdoor environments
(Fig. \ref{fig:map_compression}), 
as well as windows, ceilings, and walls in indoor environments. Thanks to the compact representation,
the proposed compact map can be generated from various HD maps of man-made environments.
Also, the map size can be further reduced for on-board and large-scale deployment.

Visual localization, as a low-cost and light-weight solution, has attracted significant study these years, 
especially the monocular solutions.
However, a monocular camera lacks direct range measurement, 
and the traditional point features are easily affected by perspective and illumination changes.
To overcome this problem, edge features are used in a model-based localization system \cite{qiu2017model}. 
Because of the rapid development of learning-based perception, 
real-time object detection \cite{redmon2018yolov3} and 
semantic segmentation \cite{chen2018encoder, zhao2018icnet, yu2020bisenet}
can help improve the robustness of visual feature-based localization.
In fact, semantic features have been used in
multiple incremental and global localization systems \cite{bowman2017probabilistic,qin2020avp}.
Compared with the widely used semantic point cloud or semantic objects, the proposed
semantic edges achieve to balance both compactness and accuracy.

In this paper, we propose a visual localization solution based on a compact map
and a low-cost sensor set consists of a monocular camera and an IMU, which can be mounted on 
vehicles, robots, and hand-held devices to realize real-time global localization. With semantic perception,
the extracted edge features are marked with specific semantic labels for
robust feature association and accurate pose estimation.

We identify our contributions as follows:
\begin{itemize}
  \item We propose a light-weight map format consisting of semantic line segments and wireframes,
        which can be generated from various $3$D maps and take very little storage space.
  \item We adopt semantic edges to achieve robust feature association and potentially
        higher accuracy compared with other semantic elements.
  \item We implement the localization system in real-time to validate the localization
        performance using real-world experiments.
\end{itemize}


\section{Related work}
\label{sec:related}
Map-based visual localization is a state estimation problem about how to register visual observations to
a pre-constructed $3$D map, a great number of works about this topic have been studied in the past few years.
According to whether or not a prior camera pose is needed, the localization algorithms can be divided into
\textit{independent localization} and \textit{prior-based localization}.

\textit{Independent localization} is also known as relocalization or loop closure detection of
simultaneous localization and mapping (SLAM). An inquiry image is the only input and the absolute
camera pose is estimated. For example, it can be realized by establishing $2$D-$3$D correspondences
between the image pixels and the 3D points of the map, followed
by a Perspective-n-Point (PnP) solver \cite{detone2018superpoint, zeisl2015camera, burki2016appearance}.
However, these methods require feature descriptors to be stored in the map, resulting in large map
sizes. In practice, an image retrieval algorithm or a rough position range can also be utilized to speed up the feature match 
process. Recently, end-to-end regression networks are used to solve visual localization in small-size 
environments \cite{brachmann2017dsac, brachmann2018learning}, but they cannot generalize to new scenarios.

\textit{Prior-based localization}, on the other hand, depends on an initial camera pose for accurate camera
pose estimation. Landmarks are selected from the map according to the initial pose for feature alignment,
it is then formulated as an optimization problem by iteratively minimizing the alignment cost
\cite{lu2017monocular, xiao2020monocular, xiao2018monocular}. 
Thus, in addition to an inquiry image, a prior camera pose is needed to make the optimization
successfully converged. This method does not need image descriptors for image-level match because of
the use of the prior camera pose, and has various feature alignment approaches in addition
to descriptor-based alignment. For instance, Zuo \textit{et.al} propose to align the semi-dense point cloud
provided by a stereo camera with a prior Lidar map \cite{zuo2019visual},
and Caselitz \textit{et.al} utilize the sparse points 
generated by monocular visual bundle adjustment instead of a stereo camera \cite{caselitz2016monocular}.
The map size can be significantly reduced compared with the map used by independent localization. 
Recently, several researchers propose to use a semantic map \cite{qin2020avp} or a vector map \cite{xiao2020monocular}
as the key localization reference, and the map size can be further reduced.

The map used for localization and the visual features used for data association are highly related.
Most traditional independent localization methods opt to utilize point features such as
SIFT \cite{lowe2004distinctive} and ORB \cite{mur2017orb},
so the corresponding maps are represented with $3$D feature points and their descriptors \cite{burki2016appearance}.
The prior-based localization methods prefer to use other alignment approaches instead of descriptor-based
feature alignment. For instance, only a point cloud map is needed by using ICP (Iterative Closest Point) alignment 
\cite{zuo2019visual, caselitz2016monocular}. Besides point features, line features or edge features 
are also widely used. They are regarded to be more robust against perspective and illumination 
changes. \cite{qiu2017model} uses a textured mesh model as the reference map, and edge features could be 
extracted from the rendered images. \cite{lu2017monocular} organizes the map as a set of road markings consists
of solid and dashed lines for urban environment localization, in which the line features are
directly used for feature association. Recently, semantic features have played an important
role in robust data association, they are proved to be more stable and robust features because of the specific
semantic information attached to the features. For urban environments, semantic elements like 
lane lines, lamp poles, traffic signs are represented in a self-defined format
\cite{xiao2018monocular} or standard vector map format \cite{xiao2020monocular} for outdoor localization,
while \cite{qin2020avp} builds a semantic point cloud map to 
represent guide signs, parking lines, speed bumps in parking lots for indoor localization.
However, most map formats mentioned above are dedicatedly designed for a certain environment.

The proposed method belongs to the prior-based localization, and semantic edge features are used for
observation-map registration. We propose to use a line segments-based map to represent
the semantic edge features. Only the semantic edges can be represented by
line segments and wireframes are used for map construction and localization landmarks,
and this concise representation way is applicable to all man-made environments.

\begin{figure*}[t]
\begin{center}
\includegraphics[width=1.9\columnwidth]{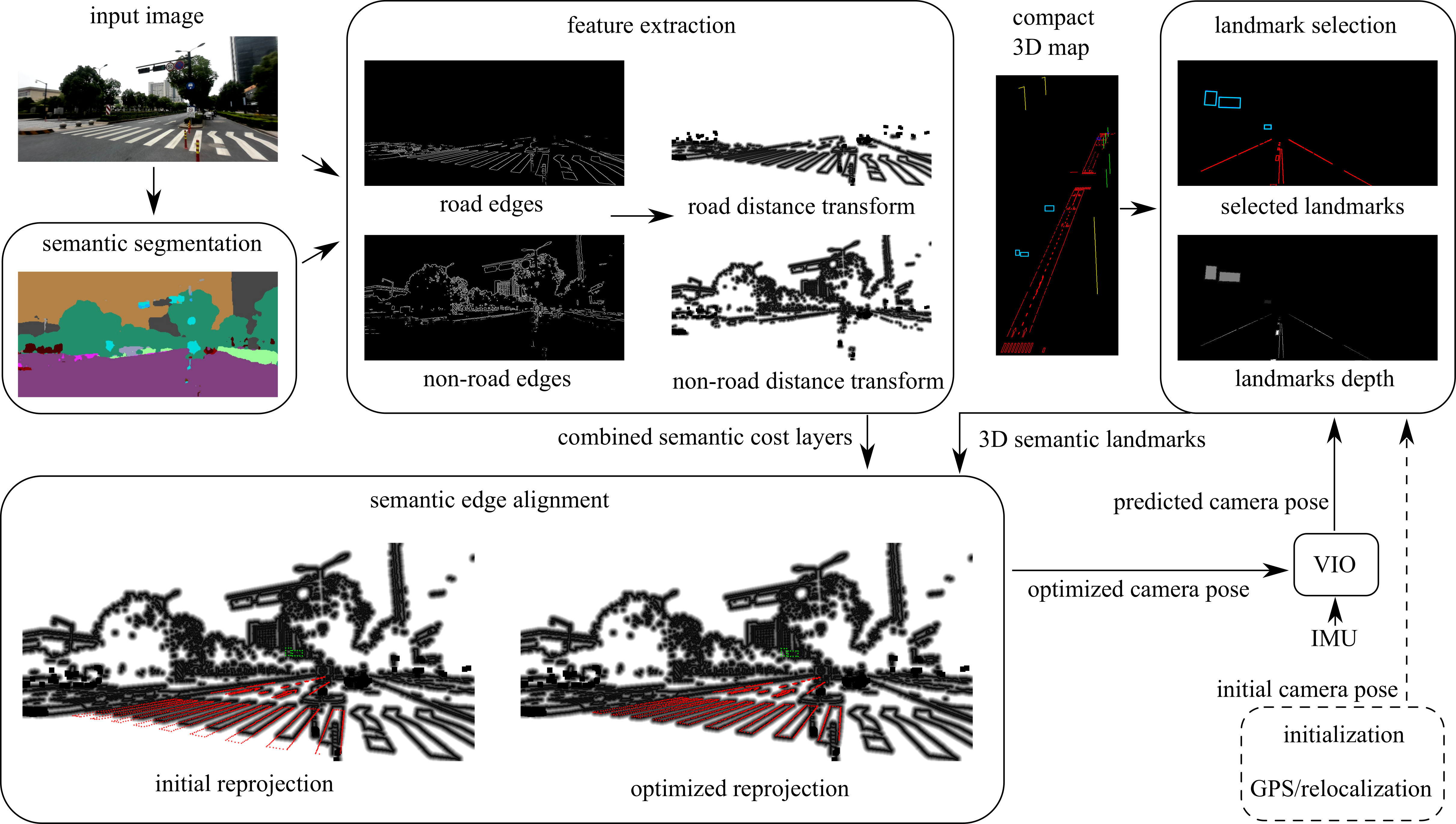}
\caption{The overall pipeline of the proposed map-based localization system. 
The whole localization system is initialized by a global reference shown in the dashed box,
all the solid boxes denote modules running in real time.
In the feature extraction module, semantic edge features
are extracted from the input image considering the segmentation results, semantic energy maps (black: low energy;
white: high energy) represented with distance transform are generated. In the landmark selection module, the landmarks
for feature alignment are selected according to the prior camera pose provided by 
an independent VIO module. In the edge alignment module, the reprojections of the 
landmarks (color points) are illustrated before and after optimization.}
\label{fig:pipeline}
\end{center}
\vspace{-5mm}
\end{figure*}

\section{System overview}
\label{sec:overview}
The proposed modular localization system includes a semantic segmentation module, a VIO
module, a landmark selection module, a feature extraction module, and a semantic edge alignment module. 
We will focus on the last three modules in this paper. For system completeness, the compact map
generation process is also briefly introduced in Section \ref{sec:methodology_map}.
Take the urban environment localization, for example, the localization pipeline is shown in Fig. \ref{fig:pipeline}.
The localization system is initialized by a
global reference such as GPS or other visual relocalization methods.
Given a captured image, the semantic segmentation module first labels the image pixels semantically, 
without loss of generality, we segment the image into two kinds of semantic areas (road and non-road).
According to the segmentation results, the potential dynamic image areas 
are masked out before further processing, and the semantic edge features are extracted from the captured
image using edge detection algorithms and the semantic segmentation results, 
the separate semantic edge images are turned into corresponding distance transforms for dense edge alignment.
Meanwhile, the current camera
pose is predicted according to the last camera pose and the odometry input from the VIO module.
With the predicted camera pose, the corresponding landmarks are selected from the pre-constructed
compact map for feature alignment.
Finally, the global camera pose is derived within an optimization framework.

For the notations used in this paper, $(\cdot)^w$ is the world frame, $(\cdot)^r$ the prior camera
frame for landmark selection, $(\cdot)^c$ the camera frame.
$\mathbf{t}_y^x$ and $\mathbf{R}_y^x$ are the $3$D translation and rotation of frame $(\cdot)^y$
with respect to frame $(\cdot)^x$.
The skew-symmetric matrix operator is denoted as $\lfloor \cdot \rfloor_\times$.
The focal lengths of the undistorted images are denoted as $f_x$ and $f_y$, respectively. The
corresponding image gradients of the distance transform are represented by $\mathrm{G}_u$ and $\mathrm{G}_v$.
$\mathrm{X}(\mathbf{p})$, $\mathrm{Y}(\mathbf{p})$, $\mathrm{Z}(\mathbf{p})$ are the $x$, $y$, $z$ values
of a $3$D point $\mathbf{p}$.
Furthermore, we define the camera projection function $\pi:\mathbb{R}^3 \longmapsto \mathbb{R}^2$ projects
the $3$D points onto $2$D images.

\section{Methodology}
\label{sec:methodology}
\subsection{Map definition \& generation}
\label{sec:methodology_map}
We define two types of landmarks, namely, a line segment and a wireframe, each
type can be attached  with multiple
semantic labels. A semantic line segment is 
represented by a semantic label and two $3$D points, and a semantic wireframe is represented 
by a semantic label and more than two $3$D points (four points for
a rectangle wireframe):
\begin{equation}
\label{equ:map}
\begin{aligned}
    \mathrm{Map} \coloneqq \{\mathbf{L}_m |_{m=0,1,\cdots}, \mathbf{W}_n |_{n=0,1,\cdots}\},
\end{aligned}
\end{equation}
in which
\begin{equation}
\label{equ:line-segments}
\begin{aligned}
    \mathbf{L}_m &= \{s, \mathbf{p}_0, \mathbf{p}_1\},\\
    \mathbf{W}_n &= \{s, \mathbf{p}_0, \mathbf{p}_1, \mathbf{p}_2, \mathbf{p}_3\},
\end{aligned}
\end{equation}
where $s$ denotes a semantic label and $\mathbf{p} \in \mathbb{R}^3$ denotes a $3$D point
in the global coordinate. The used semantic labels depend on the specific application scenarios,
for example, they will be road marks (lane lines, crosswalks, and other marks) on a road,
lamp poles on roadsides, traffic signs above a road, and skylines generated by building edges in urban
environments. Also, the map is compatible with indoor environments with different semantic
labels.

\begin{figure}[t]
\begin{center}
     \subfigure[road mark detection]{\includegraphics[width=0.9\columnwidth]{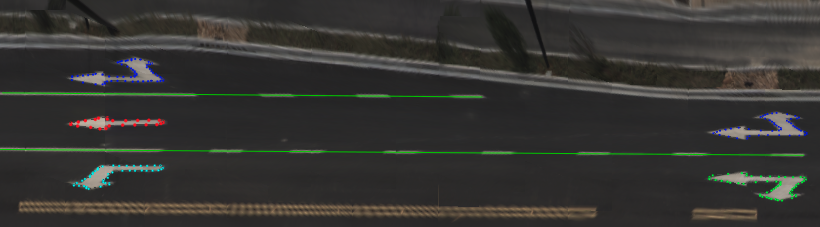}}
     \subfigure[non-road mark detection]{\includegraphics[width=0.9\columnwidth]{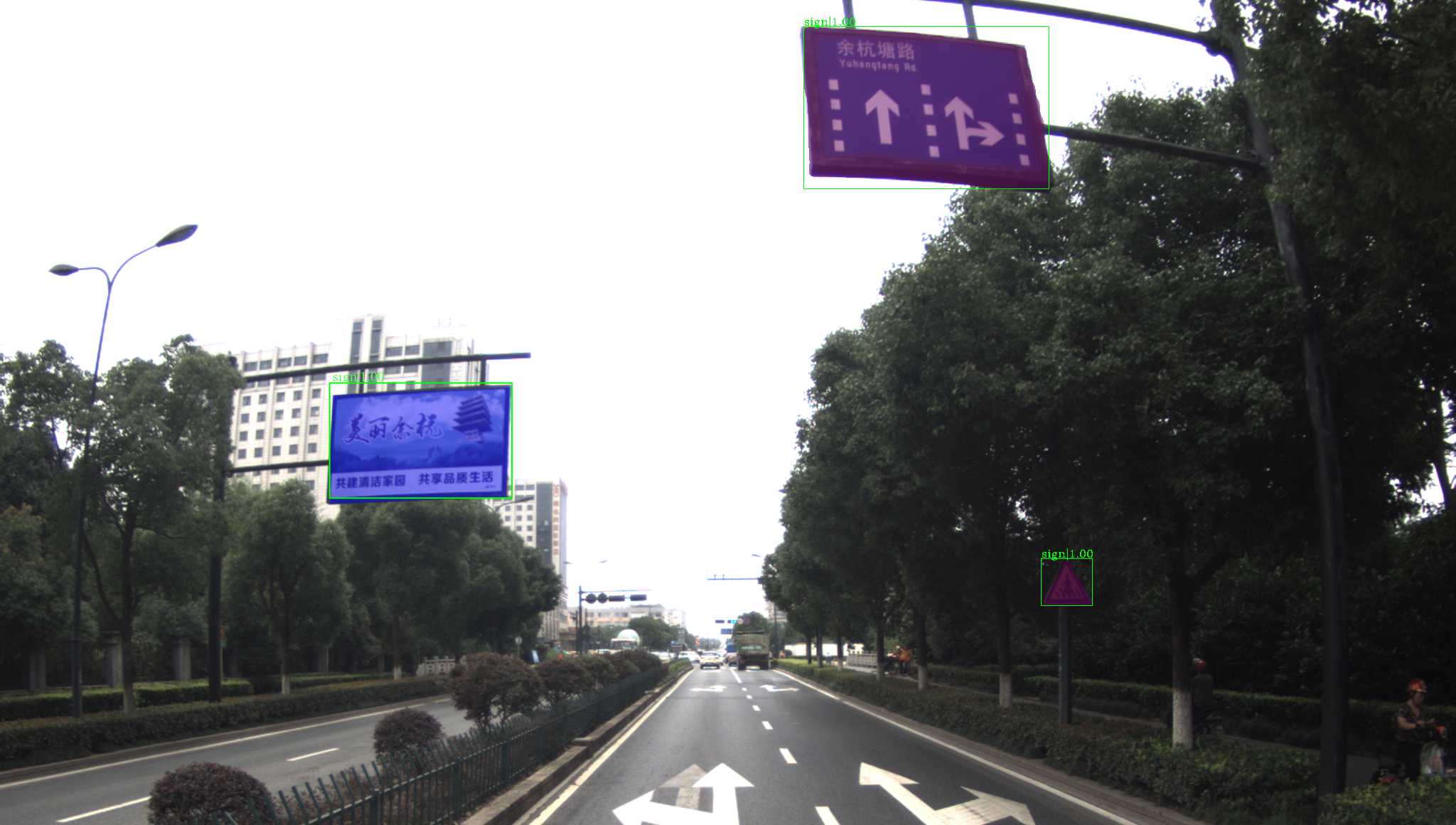}}
\end{center}
\caption{The road landmarks are detected in the
IPM (bird's eye view) image while the non-road landmarks are detected in the normal image view.}
\label{fig:landmark_detection}
\vspace{-3mm}
\end{figure}

\begin{figure}[t]
\begin{center}
\includegraphics[width=0.9\columnwidth]{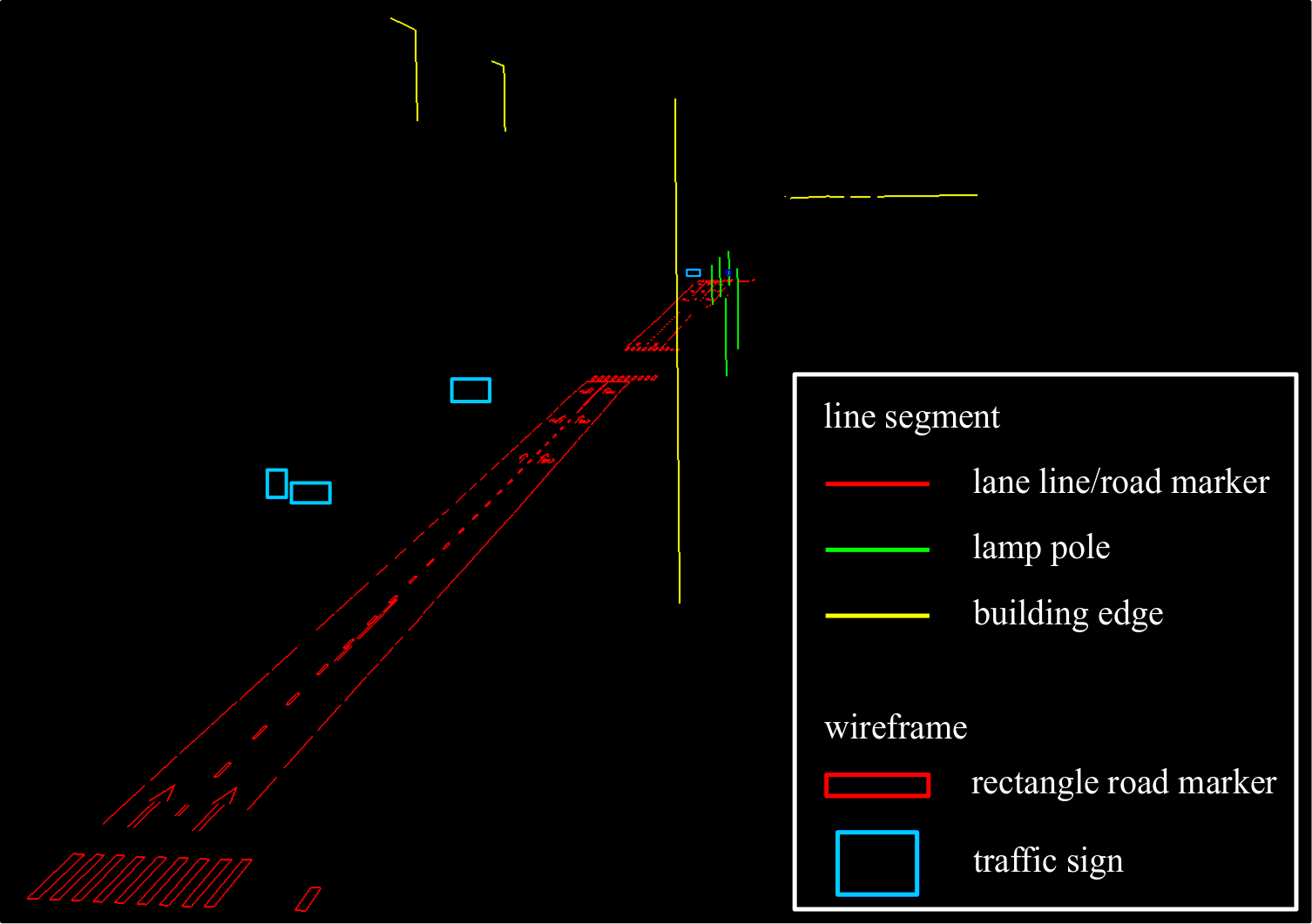}
\caption{Map generation of a sample urban environment.}
\label{fig:map_generation}
\end{center}
\vspace{-5mm}
\end{figure}

As for map generation, the compact landmarks can be converted from standard map formats or 
generated from the results of multiple mapping algorithms using various sensors.
For instance, based on an HD map constructed by visual dense mapping, the road marks can be efficiently
detected in the inverse perspective mapping (IPM) images using a segmentation
algorithm \cite{he2017mask}, as shown in Fig. \ref{fig:landmark_detection}(a). And the non-road elements
can be firstly detected in the images using a dedicated neural network and secondly projected
to the global coordinate with the help of corresponding depth information rendered from
the mapping result, as shown in Fig. \ref{fig:landmark_detection}(b). In practice, both detection
results can be refined through manual annotation for better accuracy, and the landmarks shielded
by vegetation are not kept. One sample compact map generation result is shown in Fig. \ref{fig:map_generation}.

\subsection{Landmark selection}
\label{sec:methodology_selection}
The landmarks used for prior-based localization are pre-selected
from the map according to the prior camera pose $\mathbf{R}^w_r$ $\mathbf{t}^w_r$.
In fact, the predicted camera pose is computed by fusing
the lastest reliable localization results $\mathbf{t}^w_{c_l}$, $\mathbf{R}^w_{c_l}$,
and the relative camera pose provided by the VIO module $\mathbf{t}^{c_l}_r$ $\mathbf{R}^{c_l}_r$:
\begin{equation}
\label{equ:predict_pose}
\begin{aligned}
      \mathbf{t}^w_r &= \mathbf{t}^w_{c_l} + \mathbf{R}^w_{c_l} \mathbf{t}^{c_l}_r,\\
      \mathbf{R}^w_r &= \mathbf{R}^w_{c_l} \mathbf{R}^{c_l}_r,
\end{aligned}
\end{equation}
where $c_l$ is the latest frame with reliable localization result.
In fact, only the landmarks within the prior
camera field-of-view (FOV) are selected for feature alignment. Since the landmarks may be
occluded by each other, we have to simulate the correct occlusion by considering the 
relative position of the landmarks.
For example, a traffic sign may cover the lamp poles or building edges behind it, and 
although the traffic sign is denoted by a wireframe consists of four separate line segments,
the closed rectangle area is considered for depth buffering check.
For the occlusion generated by the objects that
appeared in the real sensing stage, we design a robust feature alignment method based on the 
semantic segmentation results, which will be detailed in the following content.

In fact, the landmark selection process
is quite similar to the image rendering process in graphics: given a camera pose for
image rendering, a virtual RGB image and its depth map are generated with correct occlusion
simulation by using depth buffering, as shown in the landmark selection module of Fig. \ref{fig:pipeline}.
Thus, we also make use of depth buffering to 
generate landmarks, because some of them may partially or completely be occluded by
the landmarks in front of them.
In particular, for a lamp pole, we first expand the line segment into 
a cylinder with an averaging pole radius for further landmark rendering.
Finally, the visible landmark parts are sampled with sparse points on the
corresponding edges in the rendered images for the following edge alignment. 

\subsection{Feature extraction \& semantic edge alignment}
The key feature alignment algorithm is based on matching semantic edge features with
the corresponding semantic landmarks selected before.
Given an image for localization, the edge features are 
detected by multiple edge detection algorithms, which are further segmented with the semantic
segmentation results. 
The edges with the same semantic label are collected to formulate an 
edge image, then the corresponding semantic landmark points are reprojected
onto the edge image, the residual term of the $i$th edge point
of semantic label $s$ ($\mathbf{p}_{s_i}$ in the camera frame) 
is defined as the distance between the reprojected pixel and the nearest edge pixel:
\begin{equation}
\label{equ:reprojection_distance}
\begin{aligned}
      r_{s_i}(\mathbf{R}^w_c, \mathbf{t}^w_c)=\mathrm{min}_j \mathbf{D}
      (\pi[{\mathbf{R}^w_c}^T(\mathbf{R}^w_r\mathbf{p}_{s_i} + \mathbf{t}^w_r - \mathbf{t}^w_c)], \mathbf{u}_{s_j}),
\end{aligned}
\end{equation}
where $\mathbf{D}:(\mathbb{R}^2,\mathbb{R}^2) \longmapsto \mathbb{R}$ denotes the Euclidean 
distance between those points. If the $3$D edge points are preselected
then the minimization object function is exactly the definition of the distance
transform \cite{felzenszwalb2012distance}. It is generated from the corresponding edge image,
as shown in the feature extraction module of Fig. \ref{fig:pipeline}.
We denote the distance transform of the edge image as $V:\mathbb{R}^2\longmapsto \mathbb{R}$.
The residual term of one specific edge point becomes
\begin{equation}
\label{equ:distance_transform}
\begin{aligned}
      r_{s_i}(\mathbf{R}^w_c, \mathbf{t}^w_c)=\mathbf{V}
      (\pi[{\mathbf{R}^w_c}^T(\mathbf{R}^w_r\mathbf{p}_{s_i} + \mathbf{t}^w_r - \mathbf{t}^w_c)]).
\end{aligned}
\end{equation}

The overall energy term is formulated by accumulating all the edge point residuals
with all the semantic labels:
\begin{equation}
\label{equ:residual_term}
\begin{aligned}
      e(\mathbf{R}^w_c, \mathbf{t}^w_c) = \sum_s \sum_i (r_{s_i}(\mathbf{R}^w_c, \mathbf{t}^w_c))^2.
\end{aligned}
\end{equation}

Finally, the optimal camera pose estimation $\mathbf{R}^w_c$ and $\mathbf{t}^w_c$ can be derived by minimizing the overall
energy term. To be specific, we use the minimal representation, 
namely $\mathbf{\xi} \in \mathrm{se}(3)$ instead of $\mathbf{R}^w_c$ and $\mathbf{t}^w_c$ for optimization implementation,
and we apply the Gaussian Newton method to solve this optimization problem: 
\begin{equation}
\label{equ:gaussian_newton}
\begin{aligned}
      \mathbf{J}^T\mathbf{J}\delta \mathbf{\xi}=-\mathbf{J}^Tr(0),
\end{aligned}
\end{equation}
where $\mathbf{J}$ is the stacked matrix of all $\mathbf{J}_{s_i}$ pixel-wise Jacobians with a specific semantic label.

The key Jacobian with regard to the error state $\delta \mathbf{\xi}$ is computed as follows:
\begin{equation}
	\mathbf{J}_{s_i}=\frac{\partial \mathbf{r}_{s_i}}{\partial \mathbf{p}_{s_i}}\frac{\partial \mathbf{p}_{s_i}}{\partial \delta \mathbf{\xi}},
\end{equation}
where
\begin{align*}
	\frac{\partial \mathbf{r}_{s_i}}{\partial \mathbf{p}_{s_i}} & = \begin{bmatrix} 
	\mathrm{G}_uf_x/\mathrm{Z}(\mathbf{p}_{s_i}) \\
	\mathrm{G}_vf_y/\mathrm{Z}(\mathbf{p}_{s_i}) \\
	- \mathrm{G}_uf_x\mathrm{X}(\mathbf{p}_{s_i})/\mathrm{Z}(\mathbf{p}_{s_i})^2 - \mathrm{G}_vf_y\mathrm{Y}(\mathbf{p}_{s_i})/\mathrm{Z}(\mathbf{p}_{s_i})^2
	\end{bmatrix},
\end{align*}
$$\frac{\partial \mathbf{p}_{s_i}}{\partial \delta \mathbf{\xi}}
=[~{-\mathbf{R}^w_c}^T ~|~ \lfloor {\mathbf{R}^w_c}^T(\mathbf{R}^w_r \mathbf{p}_{s_i} + \mathbf{t}^w_r - \mathbf{t}^w_c) \rfloor_\times].$$
The camera pose $\mathbf{\xi}$ is updated by $\mathbf{\xi} = \mathrm{log}(\mathrm{exp}(\mathbf{\xi})\mathrm{exp}(\delta \mathbf{\xi}))$,
then the camera pose $\mathbf{R}^w_c$ and $\mathbf{t}^w_c$ can be recovered from it if the optimization converges.

To further evaluate the localization results, we check the consistency of the estimated camera pose and the 
predicted camera pose, then drop the ones with large pose inconsistency. Also, we check the
final averaging reprojection error, namely, the energy function value
at the final iteration over the number of edge pixels, the estimations with large reprojection
errors are also dropped. With the VIO module providing relative camera pose, the latest camera
pose can be correctly predicted by skipping the dropped frames (Equation \eqref{equ:predict_pose}).
The proposed localization method is drift-free almost all the time,
as long as the detected edge features and the selected landmarks are enough for pose estimation.

\begin{figure}[t]
\begin{center}
\includegraphics[width=0.6\columnwidth]{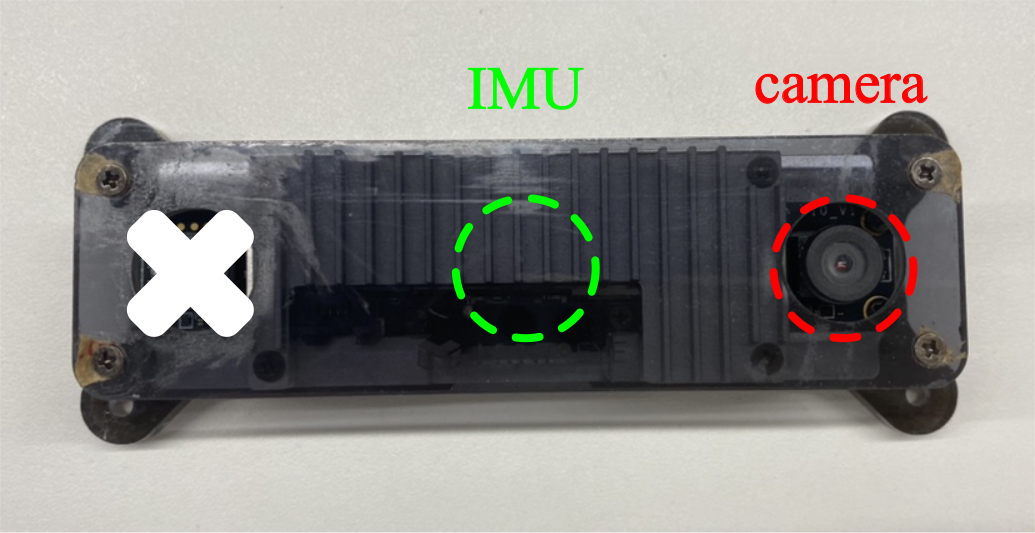}
\caption{The sensor set used for experiments. Only one camera and the inside IMU are
used for the map-based monocular localization.}
\label{fig:sensor_set}
\end{center}
\vspace{-5mm}
\end{figure}

\begin{figure}[t]
\begin{center}
\includegraphics[width=0.98\columnwidth]{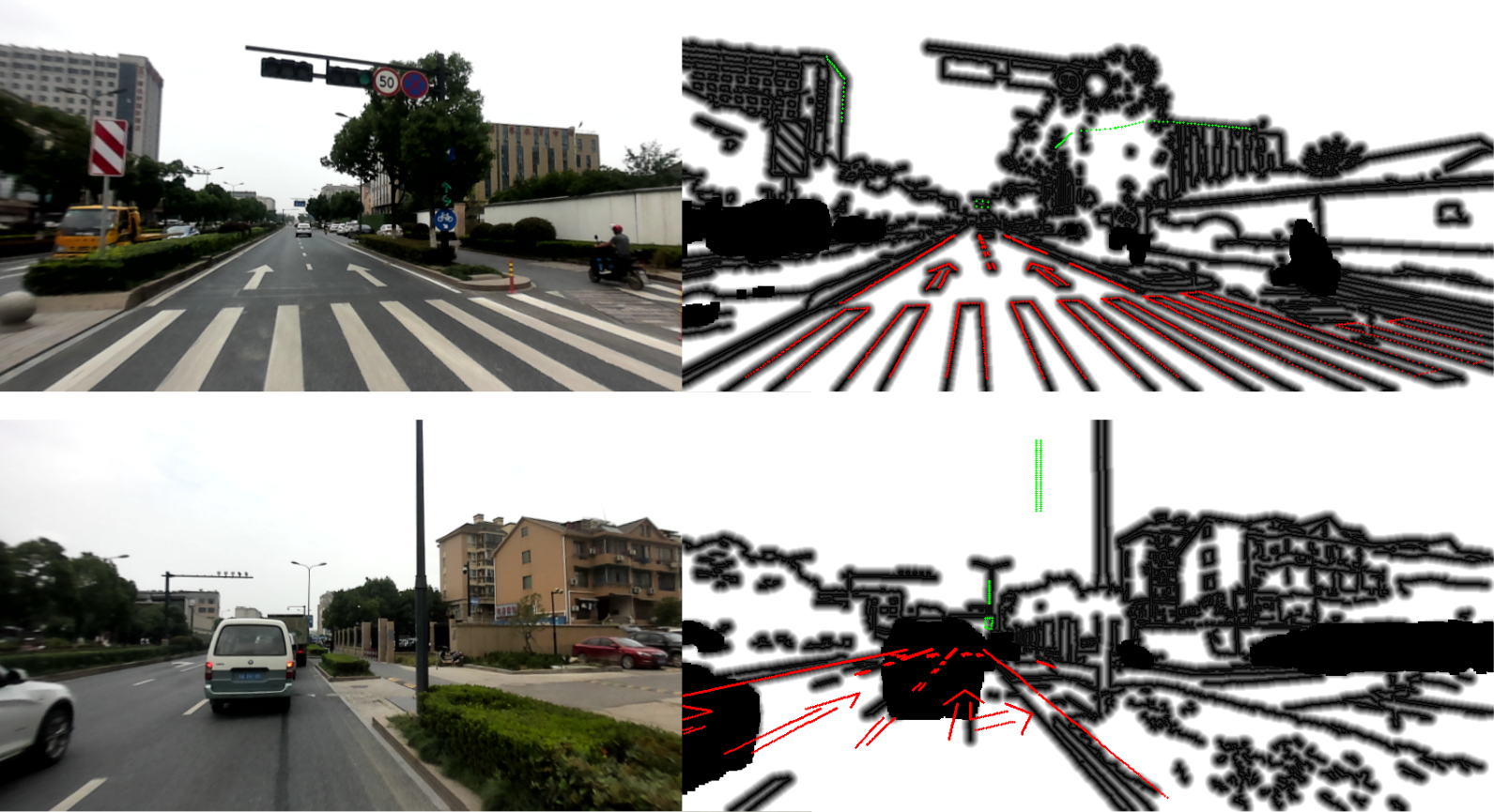}
\caption{Reprojections of selected $3$D landmarks onto the captured images after optimization
(the road and non-road distance transforms are combined, all the non-road landmarks are shown in green).
First row: the reprojected landmarks match well with the detected edges, which means the optimization converges.
Second row: the reprojected landmarks poorly match the detected edges, which means the optimization fails.}
\label{fig:reprojection}
\end{center}
\vspace{-5mm}
\end{figure}

\begin{table}[t]
	\centering
	\begin{threeparttable}
		\caption{Timing statistics.}
		\label{table:timing}
		\begin{tabular}[h]{ccc}
			\toprule
			module & time & thread\\
                  \midrule
                  semantic segmentation & $49$ ms & 1\\
                  VIO & $34$ ms & 2\\
                  landmark selection & $8$ ms & 3\\
                  feature extraction \& alignment & $37$ ms & 4\\
			\bottomrule
		\end{tabular}
	\end{threeparttable}
	\vspace{-3mm}
\end{table}

\section{Experimental results}
\label{sec:experiments}
\subsection{Implementation details}
The compact maps used for experiments are generated from a dense point cloud map
constructed by a high-end device including five industry cameras,
one high-precision SINS and one wheel odometry mounted on a data acquisition car.
On the other hand,
the sensor set used for localization includes the left monocular camera of 
an MYNT EYE camera which captures $640$ by $400$ images at $20$ Hz
and the internal IMU runs at $200$Hz, as shown in Fig. \ref{fig:sensor_set}. 
The intrinsic parameters of the camera and the extrinsic parameters between the 
camera and the IMU are calibrated in advance. The localization system
is initialized with a visual relocalization method based on SuperPoint \cite{detone2018superpoint}.
VINS-Mono \cite{qin2018vins} is used for monocular VIO implementation.
DeepLabv3+ \cite{chen2018encoder} with xception \cite{chollet2017xception} model is 
used for semantic segmentation in urban environments, resulting road and non-road areas
for further feature extraction. 
The whole localization system runs real-time on a desktop computer with an i$7$-$8700$K CPU
and a GeForce GTX $1080$ Ti graphics card, the detailed timing statistics are
shown in Table \ref{table:timing}.

\begin{table}[b]
	\centering
	\begin{threeparttable}
		\caption{Map statistics.}
		\label{table:map}
		\begin{tabular}[h]{c|cccc}
			\toprule
			\multicolumn{2}{c}{} & trial $1$ & trial $2$ & trial $3$\\
                  \midrule
                  \multicolumn{2}{c}{landmarks number} & 419 & 144 & 74\\
                  \midrule
                  \multirow{3}{*}{line segments} & lane line & 315 & 85 & 40\\
                  & lamp pole & 5 & 3 & 5\\
                  & building edge & 13 & 7 & 0\\
                  \midrule
                  \multirow{2}{*}{wireframes} & rectangle mark & 81 & 46 & 26\\
                  & traffic sign & 5 & 3 & 3\\
                  \midrule
                  \multicolumn{2}{c}{road length} & $303$ m & $171$ m & $146$ m\\
                  \multicolumn{2}{c}{original map size} & $220$MB & $184$MB & $69.4$MB\\
                  \multicolumn{2}{c}{compact map size} & $\mathbf{25.2}$\textbf{KB} & $\mathbf{9.2}$\textbf{KB} & $\mathbf{4.9}\textbf{KB}$\\
                  \multicolumn{2}{c}{compression factor} & $8.9$K & $20.5$K & $14.5$K\\
			\bottomrule
		\end{tabular}
	\end{threeparttable}
	\vspace{-3mm}
\end{table}

\subsection{Localization evaluation}
To vividly present the edge alignment results, we reproject the
landmarks onto the composed distance transform that combines all the
semantic layers for visualization, as shown in the semantic edge alignment module of Fig. \ref{fig:pipeline}.
Normally, the reprojection landmarks (color points) are located in the low energy areas (black areas of 
the distance transform) after optimization.
In other words, we can intuitively 
judge the localization results through the reprojection images.

The edge-based feature alignment is robust against illumination or appearance changes, 
and the usage of separated semantic feature layers further improves the convergence and accuracy.
But there still exist several failure cases when the selected landmarks from the map are too
few, or when severe occlusion occurs.
For example, two reprojection images after optimization
are shown in Fig. \ref{fig:reprojection}. 
If the landmarks are partially occluded, the rest landmarks still lead to correct pose estimation.
However, if the landmarks are mostly occluded by other vehicles, the localization may fail
to converge because the camera pose is not fully observable.


\begin{table}[b]
	\centering
	\begin{threeparttable}
		\caption{RMSE of pose estimations.}
		\label{table:comparison}
		\begin{tabular}[h]{ccccccccc}
			\toprule
			\multirow{2}{*}{trial} & x & y & z &\multicolumn{1}{c|}{norm} & yaw & pitch & roll & angle\\
			 & (m) & (m) & (m) & \multicolumn{1}{c|}{(m)} & ($^\circ$) & ($^\circ$) & ($^\circ$) & ($^\circ$)\\
			\midrule
			1 & 0.15 & 0.21 & 0.10 &\multicolumn{1}{c|}{0.29} & 0.19 & 0.31 & 0.15 & 0.46\\
			2 & 0.23 & 0.12 & 0.12 &\multicolumn{1}{c|}{0.29} & 0.20 & 0.41 & 0.25 & 0.52\\
			3 & 0.10 & 0.22 & 0.09 &\multicolumn{1}{c|}{0.26} & 0.12 & 0.17 & 0.20 & 0.29\\
			\bottomrule
		\end{tabular}
	\end{threeparttable}
	\vspace{-3mm}
\end{table}

To evaluate the localization accuracy of the proposed method,
we collect the localization data together with the high-end device which
are all rigidly mounted on a data acquisition vehicle (the extrinsic parameters
are calibrated in advance).
The corresponding localization results are used as the ground truth
for accuracy evaluation.
There is nothing particular about the roads we select for experiments,
pedestrians, riders, and many other vehicles are involved in the data acquisition process.
Due to the semantic segmentation module, all potential dynamic objects are
masked out from the camera view, the edge features are also marked for robust
feature association. In the meantime, the VIO module keeps
providing accurate relative pose prediction. As a result, the map-based
localization survives from several occlusion situations.

Three trials of experimental data with a total length of $620$m are collected together
with the ground truth.
The statistics of the compact maps used for localization
are shown in Table \ref{table:map}, the map of trial $1$ has a relatively 
larger landmark density.
Importantly, the map sizes are significantly reduced (compression factor $=$
original map size$/$compact map size) with the compact map representation,
which is beneficial to onboard systems and large-scale deployment.

\begin{figure}[t]
\begin{center}
      \includegraphics[width=0.95\columnwidth]{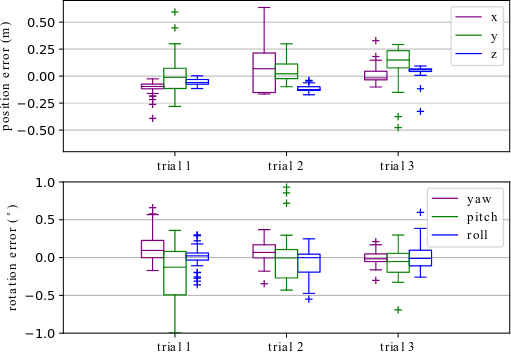}
\end{center}
\caption{Boxplot of the pose estimation error.}
\label{fig:comparison}
\vspace{-3mm}
\end{figure}

The coordinate of the localization results has been unified to the world coordinate of the ground truth
since the compact map is generated from the dense mapping result of the high-end device.
We use Z-Y-X Euler angles for rotation representation.
By considering the pre-calibrated extrinsic parameters, the pose estimation errors compared
with the ground truth are shown in Fig. \ref{fig:comparison}.
The detailed numerical results evaluated with RMSE are shown in Table \ref{table:comparison}.
Due to the use of semantic edges instead of semantic objects, the proposed method
achieves the position accuracy within $0.29$m and the rotation accuracy within $0.52^\circ$,
which satisfy the lane-level accuracy requirement of autonomous driving.
Moreover, as shown in Table \ref{table:comparison_sota}, we compare the proposed algorithm
with other map-based algorithms using their reported localization accuracy, which shows
the proposed method achieves the best performance level among the relevant works.
More localization details are demonstrated in the attached video.


\begin{table}[b]
	\centering
	\begin{threeparttable}
		\caption{Comparison of other methods.}
		\label{table:comparison_sota}
		\begin{tabular}[h]{ccccc}
			\toprule
      \multirow{2}{*}{methods} & \multirow{2}{*}{sensors} & \multirow{2}{*}{maps} & position & rotation \\
       & & & accuracy & accuracy \\
      \midrule
      \cite{caselitz2016monocular} & camera & point cloud & 0.30m & - \\
      \cite{stenborg2018long} & stereo camera & point cloud & 0.60m & - \\
      \cite{xiao2018monocular} & camera & line+point & 0.35m & 1.09$^\circ$ \\ 
      \cite{xiao2020monocular} & camera & vector map & 0.29m & - \\ 
      \textbf{ours} & camera-IMU & line segment & \textbf{0.29}m & \textbf{0.52}$^\circ$ \\ 
			\bottomrule
		\end{tabular}
	\end{threeparttable}
	\vspace{-3mm}
\end{table}

%

\subsection{Long-term localization}
In order to evaluate the long-term localization performance using the same compact map,
we also collect the sensor data for localization four months later on the same road. The 
high-end device is not used this time, but the localization results can still be 
judged by checking the reprojection landmarks in the captured images with the estimated
camera poses.
For example, the day for mapping is a cloudy day and the road surface is dry, while
the day for localization evaluation is a sunny day and the road surface is wet.
Because of the robustness of the semantic edge features we use, the captured images can
still be localized by only using the compact map constructed four months ago, more details are
demonstrated in the attached video.


\section{Conclusion and future work}
\label{sec:conclusion}
In this paper, we propose a global localization system using monocular vision and inertial measurement
based on a self-defined $3$D compact map. The map consists of two types of landmarks: a line segment and 
a wireframe, each type with multiple semantic labels. An urban environment is used to illustrate
methodology and evaluation. Moreover, thanks to the generalization of the defined map format,
it can easily adapt to
indoor environments or any man-made environments using specific semantic labels.
In addition, the semantic edge features used for feature alignment
are robust against perspective and illumination changes, resulting in long-term localization under the condition of
complicated appearance changes.
In the future,
we will generate the compact map from other map formats and
implement the localization system in other man-made environments. 

%
%

\bibliographystyle{IEEEtran}
\balance
\bibliography{localization.bib}

\begin{thebibliography}{10}
\providecommand{\url}[1]{#1}
\csname url@samestyle\endcsname
\providecommand{\newblock}{\relax}
\providecommand{\bibinfo}[2]{#2}
\providecommand{\BIBentrySTDinterwordspacing}{\spaceskip=0pt\relax}
\providecommand{\BIBentryALTinterwordstretchfactor}{4}
\providecommand{\BIBentryALTinterwordspacing}{\spaceskip=\fontdimen2\font plus
\BIBentryALTinterwordstretchfactor\fontdimen3\font minus
  \fontdimen4\font\relax}
\providecommand{\BIBforeignlanguage}[2]{{%
\expandafter\ifx\csname l@#1\endcsname\relax
\typeout{** WARNING: IEEEtran.bst: No hyphenation pattern has been}%
\typeout{** loaded for the language `#1'. Using the pattern for}%
\typeout{** the default language instead.}%
\else
\language=\csname l@#1\endcsname
\fi
#2}}
\providecommand{\BIBdecl}{\relax}
\BIBdecl

\bibitem{wolcott2014visual}
R.~W. Wolcott and R.~M. Eustice, ``Visual localization within lidar maps for
  automated urban driving,'' in \emph{2014 IEEE/RSJ International Conference on
  Intelligent Robots and Systems}.\hskip 1em plus 0.5em minus 0.4em\relax IEEE,
  2014, pp. 176--183.

\bibitem{burki2016appearance}
M.~B{\"u}rki, I.~Gilitschenski, E.~Stumm, R.~Siegwart, and J.~Nieto,
  ``Appearance-based landmark selection for efficient long-term visual
  localization,'' in \emph{2016 IEEE/RSJ International Conference on
  Intelligent Robots and Systems (IROS)}.\hskip 1em plus 0.5em minus
  0.4em\relax IEEE, 2016, pp. 4137--4143.

\bibitem{qiu2017model}
K.~Qiu, T.~Liu, and S.~Shen, ``Model-based global localization for aerial
  robots using edge alignment,'' \emph{IEEE Robotics and Automation Letters},
  vol.~2, no.~3, pp. 1256--1263, 2017.

\bibitem{redmon2018yolov3}
J.~Redmon and A.~Farhadi, ``Yolov3: An incremental improvement,'' \emph{arXiv
  preprint arXiv:1804.02767}, 2018.

\bibitem{chen2018encoder}
L.-C. Chen, Y.~Zhu, G.~Papandreou, F.~Schroff, and H.~Adam, ``Encoder-decoder
  with atrous separable convolution for semantic image segmentation,'' in
  \emph{Proceedings of the European conference on computer vision (ECCV)},
  2018, pp. 801--818.

\bibitem{zhao2018icnet}
H.~Zhao, X.~Qi, X.~Shen, J.~Shi, and J.~Jia, ``Icnet for real-time semantic
  segmentation on high-resolution images,'' in \emph{Proceedings of the
  European Conference on Computer Vision (ECCV)}, 2018, pp. 405--420.

\bibitem{yu2020bisenet}
C.~Yu, C.~Gao, J.~Wang, G.~Yu, C.~Shen, and N.~Sang, ``Bisenet v2: Bilateral
  network with guided aggregation for real-time semantic segmentation,''
  \emph{arXiv preprint arXiv:2004.02147}, 2020.

\bibitem{bowman2017probabilistic}
S.~L. Bowman, N.~Atanasov, K.~Daniilidis, and G.~J. Pappas, ``Probabilistic
  data association for semantic slam,'' in \emph{2017 IEEE international
  conference on robotics and automation (ICRA)}.\hskip 1em plus 0.5em minus
  0.4em\relax IEEE, 2017, pp. 1722--1729.

\bibitem{qin2020avp}
T.~Qin, T.~Chen, Y.~Chen, and Q.~Su, ``Avp-slam: Semantic visual mapping and
  localization for autonomous vehicles in the parking lot,'' 2020.

\bibitem{detone2018superpoint}
D.~DeTone, T.~Malisiewicz, and A.~Rabinovich, ``Superpoint: Self-supervised
  interest point detection and description,'' in \emph{Proceedings of the IEEE
  Conference on Computer Vision and Pattern Recognition Workshops}, 2018, pp.
  224--236.

\bibitem{zeisl2015camera}
B.~Zeisl, T.~Sattler, and M.~Pollefeys, ``Camera pose voting for large-scale
  image-based localization,'' in \emph{Proceedings of the IEEE International
  Conference on Computer Vision}, 2015, pp. 2704--2712.

\bibitem{brachmann2017dsac}
E.~Brachmann, A.~Krull, S.~Nowozin, J.~Shotton, F.~Michel, S.~Gumhold, and
  C.~Rother, ``Dsac-differentiable ransac for camera localization,'' in
  \emph{Proceedings of the IEEE Conference on Computer Vision and Pattern
  Recognition}, 2017, pp. 6684--6692.

\bibitem{brachmann2018learning}
E.~Brachmann and C.~Rother, ``Learning less is more-6d camera localization via
  3d surface regression,'' in \emph{Proceedings of the IEEE Conference on
  Computer Vision and Pattern Recognition}, 2018, pp. 4654--4662.

\bibitem{lu2017monocular}
Y.~Lu, J.~Huang, Y.-T. Chen, and B.~Heisele, ``Monocular localization in urban
  environments using road markings,'' in \emph{2017 IEEE Intelligent Vehicles
  Symposium (IV)}.\hskip 1em plus 0.5em minus 0.4em\relax IEEE, 2017, pp.
  468--474.

\bibitem{xiao2020monocular}
Z.~Xiao, D.~Yang, T.~Wen, K.~Jiang, and R.~Yan, ``Monocular localization with
  vector hd map (mlvhm): A low-cost method for commercial ivs,''
  \emph{Sensors}, vol.~20, no.~7, p. 1870, 2020.

\bibitem{xiao2018monocular}
Z.~Xiao, K.~Jiang, S.~Xie, T.~Wen, C.~Yu, and D.~Yang, ``Monocular vehicle
  self-localization method based on compact semantic map,'' in \emph{2018 21st
  International Conference on Intelligent Transportation Systems (ITSC)}.\hskip
  1em plus 0.5em minus 0.4em\relax IEEE, 2018, pp. 3083--3090.

\bibitem{zuo2019visual}
X.~Zuo, P.~Geneva, Y.~Yang, W.~Ye, Y.~Liu, and G.~Huang, ``Visual-inertial
  localization with prior lidar map constraints,'' \emph{IEEE Robotics and
  Automation Letters}, vol.~4, no.~4, pp. 3394--3401, 2019.

\bibitem{caselitz2016monocular}
T.~Caselitz, B.~Steder, M.~Ruhnke, and W.~Burgard, ``Monocular camera
  localization in 3d lidar maps,'' in \emph{2016 IEEE/RSJ International
  Conference on Intelligent Robots and Systems (IROS)}.\hskip 1em plus 0.5em
  minus 0.4em\relax IEEE, 2016, pp. 1926--1931.

\bibitem{lowe2004distinctive}
D.~G. Lowe, ``Distinctive image features from scale-invariant keypoints,''
  \emph{International journal of computer vision}, vol.~60, no.~2, pp. 91--110,
  2004.

\bibitem{mur2017orb}
R.~Mur-Artal and J.~D. Tard{\'o}s, ``Orb-slam2: An open-source slam system for
  monocular, stereo, and rgb-d cameras,'' \emph{IEEE Transactions on Robotics},
  vol.~33, no.~5, pp. 1255--1262, 2017.

\bibitem{he2017mask}
K.~He, G.~Gkioxari, P.~Doll{\'a}r, and R.~Girshick, ``Mask r-cnn,'' in
  \emph{Proceedings of the IEEE international conference on computer vision},
  2017, pp. 2961--2969.

\bibitem{felzenszwalb2012distance}
P.~F. Felzenszwalb and D.~P. Huttenlocher, ``Distance transforms of sampled
  functions,'' \emph{Theory of computing}, vol.~8, no.~1, pp. 415--428, 2012.

\bibitem{qin2018vins}
T.~Qin, P.~Li, and S.~Shen, ``Vins-mono: A robust and versatile monocular
  visual-inertial state estimator,'' \emph{IEEE Transactions on Robotics},
  no.~99, pp. 1--17, 2018.

\bibitem{chollet2017xception}
F.~Chollet, ``Xception: Deep learning with depthwise separable convolutions,''
  in \emph{Proceedings of the IEEE conference on computer vision and pattern
  recognition}, 2017, pp. 1251--1258.

\bibitem{stenborg2018long}
E.~Stenborg, C.~Toft, and L.~Hammarstrand, ``Long-term visual localization
  using semantically segmented images,'' in \emph{2018 IEEE International
  Conference on Robotics and Automation (ICRA)}.\hskip 1em plus 0.5em minus
  0.4em\relax IEEE, 2018, pp. 6484--6490.

\end{thebibliography}
\end{document}